\documentclass[letterpaper]{article} 
\usepackage{aaai24}  
\usepackage{times}  
\usepackage{helvet}  
\usepackage{courier}  
\usepackage[hyphens]{url}  
\usepackage{graphicx} 
\usepackage{natbib}  
\usepackage{caption} 
\usepackage{algorithm}
\usepackage{algorithmic}

\usepackage{amssymb}
\usepackage{graphicx}
\usepackage{multirow}
\usepackage{color}
\usepackage{booktabs}
\usepackage{comment}
\usepackage{amsmath}
\usepackage{amssymb}
\usepackage{multirow}
\usepackage{subcaption}
\usepackage{pifont}
\usepackage{xcolor}
\usepackage{subcaption}

\newcommand{\xmark}{\ding{55}}

\def\mD{\mathcal{D}}
\def\mE{\mathcal{E}}

\def\mL{\mathcal{L}}

\def\mO{\mathcal{O}}

\def\mS{\mathcal{S}}

\def\mW{\mathcal{W}}
\def\mX{\mathcal{X}}

\def\1n{\mathbf{1}_n}
\def\0{\mathbf{0}}
\def\1{\mathbf{1}}

\newcommand{\myargmax}[1]{\mathop{\textrm{argmax}}_{#1}} 
\newcommand{\myst}{\textrm{s.t. }}

\definecolor{pink}{rgb}{0.9,0.5,0.5}
\definecolor{purple}{rgb}{0.5, 0.4, 0.8}   
\definecolor{gray}{rgb}{0.3, 0.3, 0.3}
\definecolor{mygreen}{rgb}{0.2, 0.6, 0.2}

\definecolor{greena}{rgb}{0.4, 0.5, 0.1}

\definecolor{bluea}{rgb}{0, 0.4, 0.6}

\definecolor{reda}{rgb}{0.6, 0.2, 0.1}

\newcommand{\cm}[1]{}

\newcommand{\myheading}[1]{\vspace{1ex}\noindent \textbf{#1}}

\newif\ifshowsolution
\showsolutiontrue

\ifshowsolution  

\else

\fi

\newcommand{\Fref}[1]{Fig.~\ref{#1}}
\newcommand{\Tref}[1]{Table~\ref{#1}}

\usepackage{newfloat}
\usepackage{listings}
\DeclareCaptionStyle{ruled}{labelfont=normalfont,labelsep=colon,strut=off} 
\floatstyle{ruled}
\newfloat{listing}{tb}{lst}{}
\floatname{listing}{Listing}
\title{Count What You Want: Exemplar Identification and \\ Few-shot Counting of Human Actions in the Wild}
\author{
Yifeng Huang\textsuperscript{\rm *1},\quad Duc Duy Nguyen\textsuperscript{\rm *2},\quad Lam Nguyen\textsuperscript{\rm 2}, \quad Cuong Pham\textsuperscript{\rm 2,3}, \quad Minh Hoai\textsuperscript{\rm  1,2}
}

\affiliations{
    \textsuperscript{*} Equal contribution \quad  
    \textsuperscript{\rm 1}Stony Brook University, NY, USA \quad 
    \textsuperscript{\rm 2}VinAI, Hanoi, Vietnam \\ 
    \textsuperscript{\rm 3}Posts \& Telecommunications Institute of Technology, Hanoi, Vietnam

}

\begin{document}

\maketitle

\begin{abstract}
This paper addresses the task of counting human actions of interest using sensor data from wearable devices. We propose a novel exemplar-based framework, allowing users to provide exemplars of the actions they want to count by vocalizing predefined sounds ``one'', ``two'', and ``three''. Our method first localizes temporal positions of these utterances from the audio sequence. These positions serve as the basis for identifying exemplars representing the action class of interest. A similarity map is then computed between the exemplars and the entire sensor data sequence, which is further fed into a density estimation module to generate a sequence of estimated density values. Summing these density values provides the final count. To develop and evaluate our approach, we introduce a diverse and realistic dataset consisting of real-world data from 37 subjects and 50 action categories, encompassing both sensor and audio data. The experiments on this dataset demonstrate the viability of the proposed method in counting instances of actions from new classes and subjects that were not part of the training data. On average, the discrepancy between the predicted count and the ground truth value is 7.47, significantly lower than the errors of the frequency-based and transformer-based methods. Our project, code and dataset can be found at \url{https://github.com/cvlab-stonybrook/ExRAC}.
\end{abstract}

\section{Introduction}
\label{sec:intro}

Counting human actions of interest using wearable devices is a crucial task with applications in health monitoring~(e.g.,~\citet{baghdadi2021monitoring}) and performance evaluation~(e.g.,~\citet{o2018wearable}). However, the majority of existing counters are often designed for a limited set of action categories, such as walking and a few other physical exercises. These class-specific counters (e.g.,~\citet{genovese2017smartwatch}) are incapable of handling classes beyond those they have been explicitly trained for. Consequently, relying solely on class-specific counters becomes impractical and unscalable when dealing with a diverse set of action categories. For scalability, a promising alternative to class-specific counters is class-agnostic counters, capable of tallying repetitions from any arbitrary class, as long as this class represents the dominant activity within the sensor data being analyzed. 

\begin{figure*}[t]
    \centering
    \includegraphics[width=0.80\textwidth]{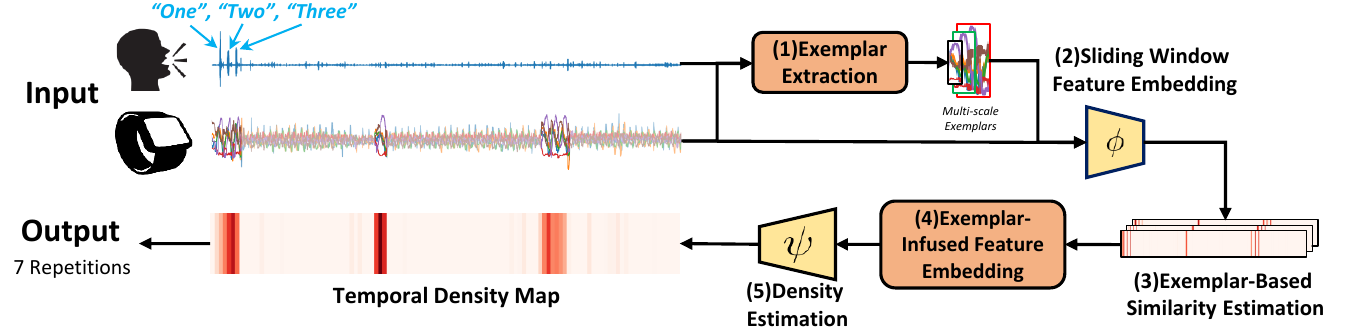}
    \vspace{-8px}
    \caption{{\bf Processing pipeline of our method}. The input consists of the sensor signal and the audio signal containing the utterances ``one,'' ``two,'' and ``three,'' corresponding to three repetitions of the action of interest. The output is the total count, obtained by summing the values of the intermediate 1D density profile. This profile is better visualized as a 2D map as shown here. This figure also shows the other processing steps, which will be explained in the forthcoming method section.}
    \label{fig:teaser}
    \vspace{-18px}
\end{figure*}

However, in many real-world scenarios, our interest might not lie in counting actions from the dominant class. For instance, in sports training and skill evaluation, the objective is often to detect specific and infrequent mistakes within the prevalent data. As illustrated in \Fref{fig:teaser}, the action of interest may occur only briefly within the entire data sequence. These factors pose significant challenges when applying existing methods effectively.

Confronting the challenge presented by real-world data, which often contains undesired actions, we propose to develop an exemplar-based counting method, where an user can provide exemplars of what they want to count. However, the development of such a method poses two significant technical challenges. Firstly, devising a convenient exemplar provision scheme is nontrivial. Secondly, once we have some exemplars, the question remains how to effectively leverage them. In this paper, we address both of these challenges to develop a novel exemplar-based counting method. 

For the first challenge, we propose an intuitive and non-intrusive approach for specifying exemplars using vocal sounds. The exemplars are conveniently provided by verbally counting out loud ``one,'' ``two,'' ``three'' at the onset of the counting process as shown in \Fref{fig:teaser}. Each utterance corresponds to one repetition. To accurately detect the positions of these counting utterances in the audio sequence, we develop an efficient algorithm that solves a constrained optimization problem with the two constraints on the temporal ordering and the temporal distance between the identified positions. Once the positions of the counting utterances are identified, we extract the exemplars from these locations. 

For the second challenge, we propose a novel model that jointly processes the exemplars with the whole data sequence as shown in \Fref{fig:teaser}. More concretely, we first generate per-window embeddings for both the exemplars and the whole data sequence. Subsequently, we compute a similarity map between the exemplar and  data sequence embeddings, using Soft-DTW~\cite{DBLP:conf/icml/CuturiB17} and correlation measures. 
This similarity map serves as the basis for generating a sequence of exemplar-infused embeddings for the data sequence. The initial embedding sequence and  the exemplar-infused embedding sequence are then fed into a density estimation module for moment-by-moment density estimation, from which the final count is obtained by summing the density values.

Realizing the importance of a good similarity measurement, we introduce a novel distance-preserving loss. This loss enforces the high-dimensional per-window embeddings to maintain local patterns, thereby preserving the similarity relationships observable in the lower-dimensional space. In addition, considering the limited training data, we propose an exemplar-based data synthesis pipeline, which can synthesize training data and improve the result significantly.

To develop and evaluate the proposed method, we have collected a  dataset named \textbf{D}iverse \textbf{W}earable \textbf{C}ounting dataset (\textbf{DWC}). This dataset comprises sensor data sequences accompanied by audio-specified exemplars collected from 37 subjects performing 50 distinct action categories. What sets this dataset apart from many existing ones is the availability of synchronized audio data with vocal sounds for specifying exemplars. Furthermore, this dataset includes instances where the action of interest may not be the predominant action within the data sequence, providing a more realistic representation of real-world scenarios.

In short, the main contributions of our paper are threefold. First, we introduce a novel strategy for using audio to specify exemplars of what needs to be counted. Second, we propose a novel counting method that utilizes exemplars, incorporating a distance-preserving loss and an exemplar-based data synthesis pipeline. Third, we introduce an unique dataset with multiple data modalities to develop a practical counting method for real-world scenarios. 

\section{Related Work}

Action counting through wearable devices is driven by its diverse range of applications in health monitoring~\cite{baghdadi2021monitoring, lee2015heart, nam2016sleep, hatamie2020textile, ramachandran2022microfluidic, patel2010novel}, sports training~\cite{chang2007tracking, o2018wearable, kranz2013mobile, ding2015femo}, and industrial contexts~\cite{kong2019industrial, stiefmeier2008wearable}. 
Existing counting methodologies have predominantly focused on particular action categories, such as physical exercises~\cite{genovese2017smartwatch, kupke2016development, pillai2020personalized, bian2019passive, ishii2021exersense, morris2014recofit, soro2019recognition, oh2020crowd}. This specialization restricts their adaptability, especially when faced with classes having no prior training data. Consequently, relying on class-specific counters proves inadequate and unscalable in managing the wide range of action categories encountered in real world. 

Class-agnostic counters is an alternative to class-specific counters, but they can only count repetitions from the dominant class. Earlier strategies, based on Fourier analysis or wavelet transforms~\cite{DBLP:journals/pami/CutlerD00, DBLP:conf/icpr/AzyA08, DBLP:conf/cvpr/PogalinST08, DBLP:conf/cvpr/RuniaSS18}, peak detection~\cite{DBLP:conf/wacv/ThangaliS05}, and singular value decomposition~\cite{DBLP:conf/bmvc/ChetverikovF06}, have been explored. More recently, significant attention has been directed towards repetitive action counting in videos~\cite{DBLP:conf/iccv/LevyW15, DBLP:conf/cvpr/ZhangXHH20, DBLP:conf/cvpr/Zhang0S21, DBLP:conf/cvpr/FieraruZPOS21, DBLP:journals/corr/abs-2107-13760, DBLP:conf/cvpr/HuDZLLG22, DBLP:conf/cvpr/DwibediATSZ20}. Recent works~\cite{DBLP:conf/cvpr/DwibediATSZ20, DBLP:conf/cvpr/HuDZLLG22} have achieved promising results by harnessing temporal self-similarity to count repetitive actions from the dominant class.

While exemplar-based counting is not a novel concept, our contribution stands as one of the few approaches designed for wearable devices. Notably, it marks the pioneering effort in introducing a strategy for specifying exemplars through the act of uttering and subsequently detecting predefined vocal sounds. This approach is innovative and distinct from existing works in various fields. For instance, in computer vision, there are methods that utilize exemplars for counting objects in images~\cite{DBLP:conf/bmvc/LiuZZX22, DBLP:conf/wacv/YangSHC21, DBLP:conf/cvpr/RanjanSNH21, DBLP:conf/cvpr/RanjanH22, DBLP:conf/cvpr/Shi0FL022,  DBLP:conf/accv/LuXZ18, DBLP:conf/wacv/YouYLLCL23, DBLP:conf/eccv/NguyenPNH22, m_Huang-etal-ICCV23, m_Ranjan-Hoai-ACCV22}. These methods require users to  specify exemplars by drawing bounding boxes. However, when dealing with time-series data, the natural provision of exemplars becomes non-trivial. First, the visualization and semantic parsing of sensor data pose greater challenges compared to images. Second, manually determining the temporal extents of human actions in time series is more difficult compared to delineating object bounding boxes in images. Third, for sensor-based counting, immediate results are often required, making it crucial for the process of providing and identifying exemplars to be convenient and efficient, without involving time-consuming procedures such as transmitting, visualizing, and drawing. 

\begin{figure*}[t]
    \centering
    \includegraphics[width=0.8\textwidth]{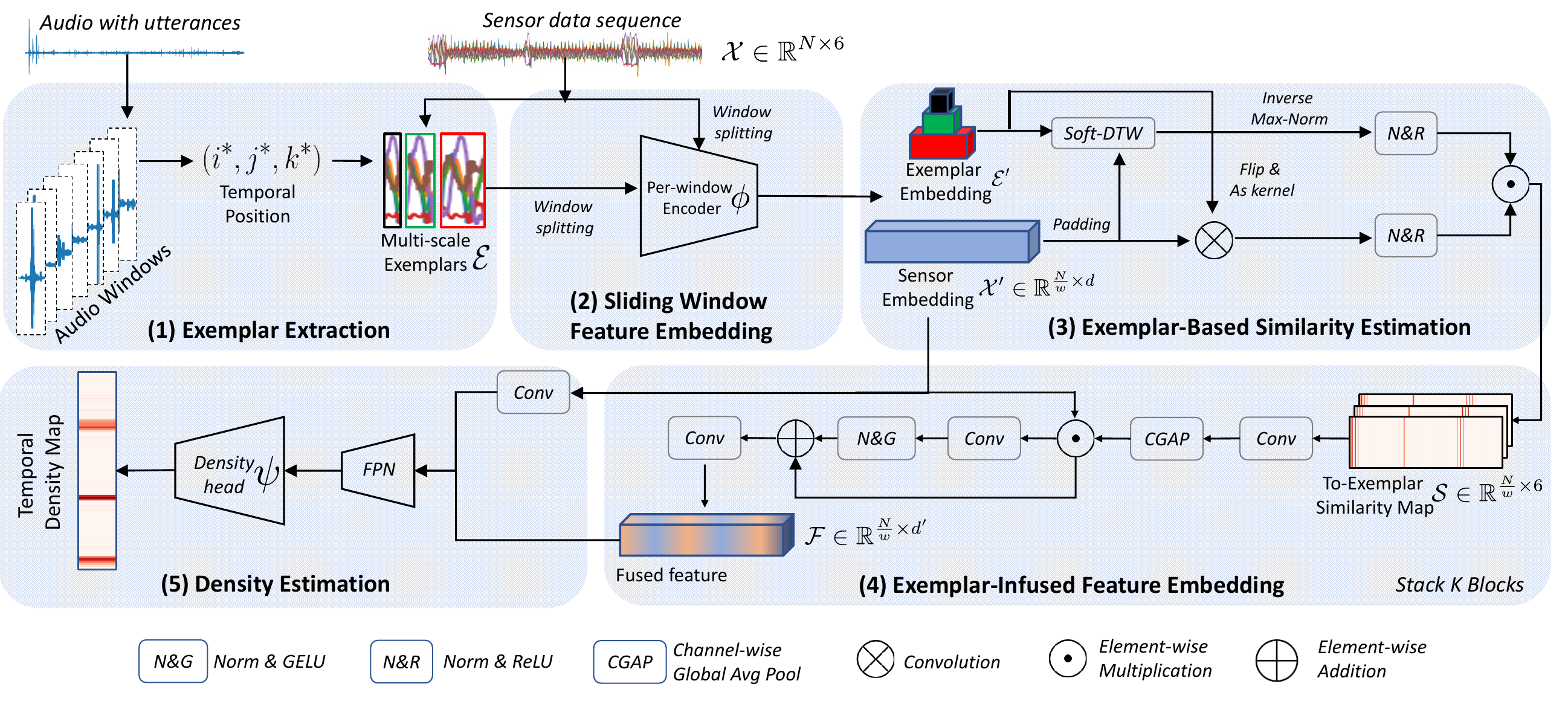}
    \vspace{-8px}
    \caption{{\bf Main steps of our method.} Our method begins with exemplar extraction, which is based on predefined utterance detection in the audio data. Following this, per-window embeddings are extracted. Subsequently, we compute the similarity between the entire sensor sequence and the exemplars, which is then used for feature fusion. Finally, the temporal density map is estimated based on the fused features and the sensor embeddings.}
    \label{fig:pipeline}
    \vspace{-20px}
\end{figure*}

\section{Proposed Approach}
\label{sec:app}

Our objective involves tallying the occurrences of a specific action class within a sequence of sensor data. Our method takes as input both the sensor data sequence and an audio sequence synchronized with it, featuring predetermined vocal sounds – one, two, three – corresponding to the initial three repetitions of the action. As such, our approach comprises two fundamental stages: first, the identification of exemplars, and subsequently, their utilization to derive the overall count. These stages are executed using five  modules, as depicted in \Fref{fig:pipeline}: (1) exemplar extraction, (2) sliding window feature embedding, (3) exemplar-based similarity estimation, (4) exemplar-infused feature embedding, and (5) density estimation. In this section, we will elucidate these five modules along with the training procedure.

\subsection{Exemplar Extraction \label{sec:app:exemplar}}
To extract the exemplars for the action class of interest, we first identify three temporal positions corresponding to the predefined vocal sounds (one, two, three) in the audio. A naive approach is to use a pre-trained classifier to greedily select the window with the highest classification score. However, this fails to exploit two critical cues: (1) temporal ordering, which requires the order of the sounds one, two, threes to be preserved, and (2) temporal proximity, which ensures that the distance between two predefined sounds is not excessively large. Considering these two properties, we formulate the temporal position detection into a constrained optimization problem as follows:
\begin{align}
i^*, j^*, k^* & = \myargmax{i,j,k} C_i^1 C_j^2C_k^3, \\ 
 \myst &  1 \leq i<j<k \leq M \ \textrm{and } k-i \leq R. 
\end{align}

Here, $i, j, k$ denote the indices of a sliding window. $C_{i}^{u}$ is the classification score for the $i^{th}$ window to be the $u^{th}$ utterance. $R$ is the upper bound for the temporal distance.

The above optimization problem can be solved efficiently using dynamic programming. We first divide the audio signal into $M$ overlapping sliding windows, each with a duration of one second and the step size being 0.1 seconds. We then compute the classification scores $(C_{i}^1, C_{i}^2, C_{i}^3)$ for each window  using a pre-trained classifier, specifically the BC\_ResNet~\cite{DBLP:conf/interspeech/KimC0S21} pretrained on Speech Command~\cite{DBLP:journals/corr/abs-1804-03209}. For every group of $R$ consecutive windows, we optimize $C_i^1 C_j^2C_k^3$ subject to the only constraint $i<j<k$ with dynamic programming. The complexity of this algorithm is $\mO(R)$, and we have to run it $M-R+1$ times for $M-R+1$ groups of $R$ consecutive windows. Thus, the overall complexity is $\mO(R(M-R+1))$. 

Let $\mathcal{X} \in \mathbb{R}^{N \times d}$ denote the sensor data sequence, with $N$ being the length and $d$ the number of sensor values at each time step ($d=6$ for data from the accelerometer and gyroscope of a smartwatch). Upon solving the above optimization problem, we obtain $i^*, j^*, k^*$, which indicate the locations of the three exemplars. To avoid noisy exemplars, we only retain the two locations with the highest classification confidence and let them be denoted as $s_1$ and $s_2$. Unfortunately, we do not know the temporal extents of the exemplars. To address this issue, we adopt a multi-scale approach as follows. For each position $s$ among the two positions $s_1$ and $s_2$, we extract three exemplar sequences corresponding to three different scales: $\mathcal{X}[s-10:s+10]$, $\mathcal{X}[s-20:s+ 20]$, and $\mathcal{X}[s-40:s+ 40]$.  With two locations and three scales, we have a total of six exemplars. This strategy enables us to count actions at various levels of granularity. 

\subsection{Sliding Window Feature Embedding}
\label{sec:app:encoder}

As sensor values at individual time steps carry limited information, we learn and use window-level sensor representation instead. To accomplish this, we partition a sensor data sequence into non-overlapping windows, with each window comprising $w$ sensor data points. We subsequently embed each window into a high-dimensional representation turning the sequence of original sensor values 
$\mX \in \mathbb{R}^{\frac{N}{w}\times d}$ into a sequence of embedding vectors $\mathcal{X}' \in\mathbb{R}^{\frac{N}{w}\times d'}$. Let $\phi$ denote this mapping, i.e., $\mathcal{X}' = \phi(\mathcal{X})$, and $\phi$ is implemented using temporal convolution. Specifically in our experiments, $w$ is set to 10, and $d'$ is set to 64. Similarly, the exemplar sequence $\mE$ is transformed into $\mE'$ using $\phi$.

\subsection{Exemplar-Based Similarity Estimation} 
\label{sec:app:similarity}
Utilizing per-window embedding, we estimate the similarity map $\mathcal{S}$ between the sensor embedding $\mathcal{X}'$ and the exemplar embedding $\mathcal{E}'$. Correlation and Dynamic Time Warping (DTW) are two widely-used methods for estimating similarity between sequential data. However, directly applying them to estimate the similarity between $\mathcal{X}'$ and $\mathcal{E}'$ is not effective because correlation is sensitive to differences in scale and offset while DTW tends to overreact to static data. To address these issues, we combine DTW and correlation to estimate the similarity as follows.

We first compute the correlation  between the whole sequence embedding and the exemplar embedding: 
$\mathcal{S}^c = ReLU(\text{Norm}(\mathcal{X}' \otimes \mathcal{E}'))$
where $\otimes$ is correlation operation with zero-padding to preserve the length of the signal (i.e., $\mS^c$ and $\mX'$ have the same length). Next, we calculate the Soft-DTW similarity~\cite{DBLP:conf/icml/CuturiB17} between the exemplar embedding and the sliding window on the whole sequence embedding. For the sliding window at location $i$, the resulting value is $\mS^d_i = \text{Soft-DTW}(\mX'[i-\frac{k}{2}, i+\frac{k}{2}], \mathcal{E}')$, where $k$ is the length of the exemplar $\mE'$. Then, $\mathcal{S}^{d}$ is fed into normalization and ReLU layers as $\mathcal{S}^{d} = ReLU(\text{Norm}(\text{Max}(\mathcal{S}^{d}) - \mathcal{S}^{d}))$. Considering that Soft-DTW estimates the distance between two samples, we transform it into a measure of similarity by taking the negative of the distance and adding the maximum value, thereby ensuring a non-negative similarity measure. The final similarity profile is obtained by computing $\mathcal{S} = \mathcal{S}^c \odot \mathcal{S}^d$, where $\odot$ denotes element-wise multiplication. Since we have two exemplars at three scales, the dimension of $\mathcal{S}$ is $\mathcal{S} \in \mathbb{R}^{\frac{N}{w}\times{6}}$.

\subsection{Exemplar-Infused Feature Embedding}
\label{sec:app:fusion}
Upon obtaining the  similarity map $\mathcal{S}$, we use it to generate a refined representation that emphasizes exemplar-related features while suppressing irrelevant features. This can be implemented with a stack of $K$ fusion blocks, and the process can be described as follows:
\begin{align}
& \mathcal{F}_0 = \mathcal{X}',\ \mathcal{S}_0 = \mathcal{S}, \\
&\mathcal{S}_i = \text{CGAP}(\text{Conv}(\mathcal{S}_{i-1})),\\
&\mathcal{F}_i = \text{Conv}(\mathcal{F}_{i-1} +\text{GELU}(\text{Norm}(\text{Conv}(\mathcal{F}_{i-1} \odot \mathcal{S}_{i-1})))). \nonumber 
\end{align}
Here, $\text{CGAP}$ is the channel-wise (among exemplars) global average pooling, and $\odot$ denotes element-wise multiplication. The final fused feature is $\mathcal{F} = \mathcal{F}_K \in\mathbb{R}^{\frac{N}{w}\times d'}$. 

\subsection{Density Estimation}
\label{sec:app:density}
The density estimation head comprises a Feature Pyramid Network (FPN) designed to extract multi-scale features and a temporal convolution counting head $\psi$ to estimate the temporal densities. We extract multi-scale features as follows:
\begin{align}
& \mathcal{F}_{s_1}, \mathcal{F}_{s_2}, \mathcal{F}_{s_3} = \text{FPN}(\mathcal{F}),\\ 
& \mathcal{X}'_{s_1}, \mathcal{X}'_{s_2}, \mathcal{X}'_{s_3} = \text{FPN}(\text{Conv}(\mathcal{X}')),
\end{align}
where $\mathcal{F}_{s_1}, \mathcal{F}_{s_2}, \mathcal{F}_{s_3}$ are multi-scale fused features from low to high, and $\mathcal{X}'_{s_1}, \mathcal{X}'_{s_2}, \mathcal{X}'_{s_3}$ are multi-scale sensor feature for the sensor embedding. Using max-pooling, $\mathcal{F}_{s_1}, \mathcal{F}_{s_2}, \mathcal{X}'_{s_1}, \mathcal{X}'_{s_2}$ are down-sampled to have the same length as $\mathcal{F}_{s_3}$ and $\mathcal{X}'_{s_3}$. All of them are then concatenated and fed into a density estimation head $\psi$, implemented with a temporal convolution network.


\subsection{Training loss}

\label{sec:app:loss}
The counting loss over the predicted temporal density map is given by the squared error of the final count, expressed as: $\mathfrak{L}_{c} = (\text{sum}(\mathcal{T}) - \hat{c})^2$, where $\hat{c}$ is the ground truth count.

The success of our method largely depends on accurately estimating similarity between the exemplars and the query data sequence. However, it's important to note that the similarity relationship within the raw data space $\mathcal{X}$ may not be fully preserved in the embedding space $\mathcal{X}'$. This is especially true when dealing with limited training data and the lack of a robust pre-trained feature extractor. Inspired by Laplacian Eigenmaps~\cite{DBLP:journals/neco/BelkinN03}, we propose to use a distance-preserving loss to encourage the per-window encoder to preserve the relationship of distance by enforcing the encoder to maintain the local patterns. We first build a $k$-nearest-neighbor graph over the raw window to represent the local pattern. To build it, we compute the adjacency matrix $\mathcal{W}$, where $\mathcal{W}_{ij} = \exp(-\frac{||\mathcal{X}_i - \mathcal{X}_j||^2}{2\sigma^2})$ represents the similarity between the $i^{th}$ window and $j^{th}$ window. Then, for each node in the graph, we retain the top $k$ nearest neighbors in the adjacency matrix ($k=150$ in our work). We compute the graph Laplacian: $\mL = \mD - \mW$, where $\mD$ is the degree matrix with $\mD_{ii} = \sum_{j}{\mathcal{W}_{ij}}$ and $\mD_{ij} = 0$ for $i\neq j$. Then the distance-preserving loss is defined as 
$\mathfrak{L}_{pl} = \mathcal{X}'^T\mathcal{L} \mathcal{X}'$. The overall training loss is: 
$\mathfrak{L}_{train} = \mathfrak{L}_{c} + \lambda\mathfrak{L}_{pl}$, 
where $\lambda$ is set to 0.01.

\begin{figure*}[t]
  \centering
  \begin{subfigure}[c]{0.23\textwidth}
    \centering
    \includegraphics[width=\textwidth]{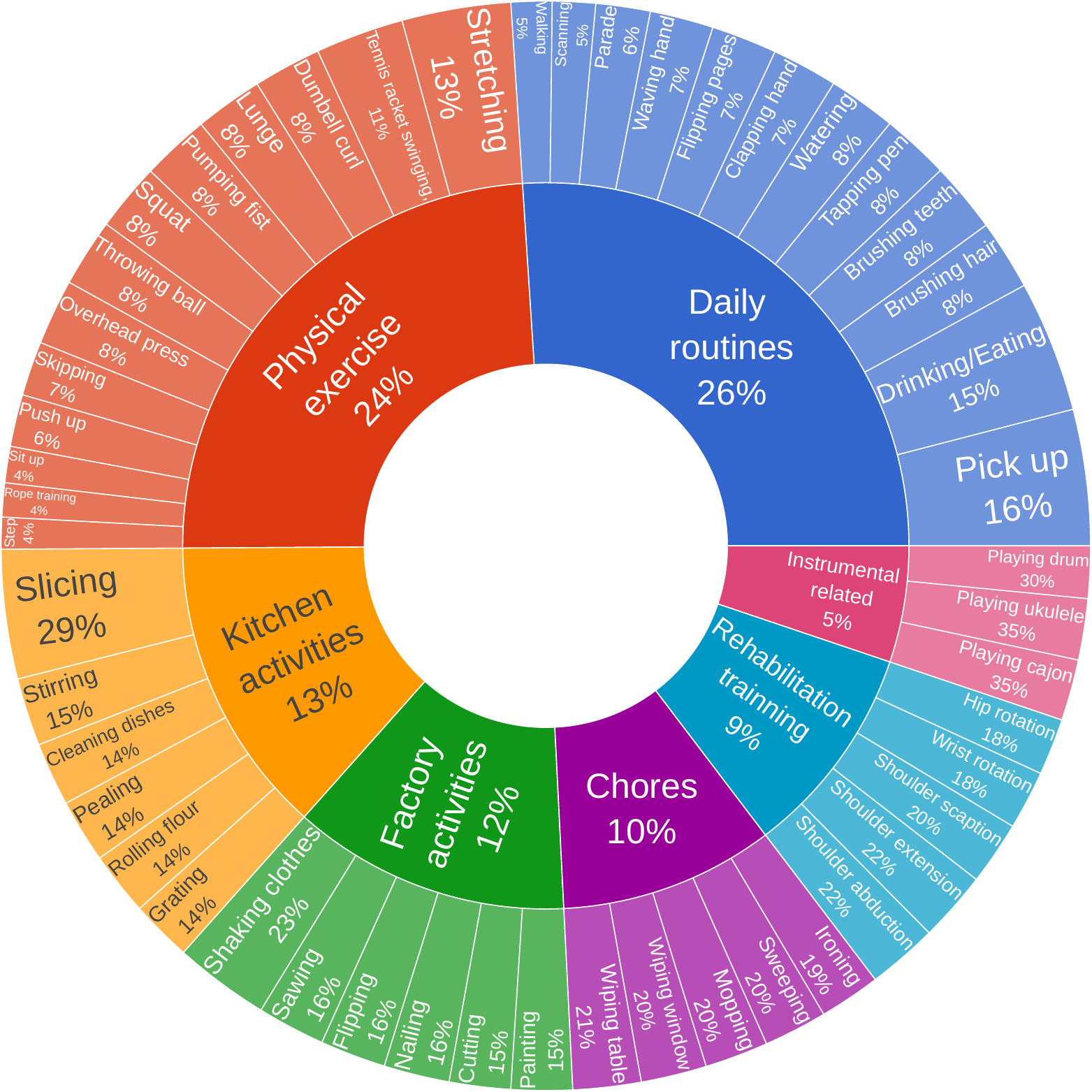}
  \end{subfigure}
  \hfill
  \begin{subfigure}[c]{0.30\textwidth}
    \centering
    \includegraphics[width=\textwidth]{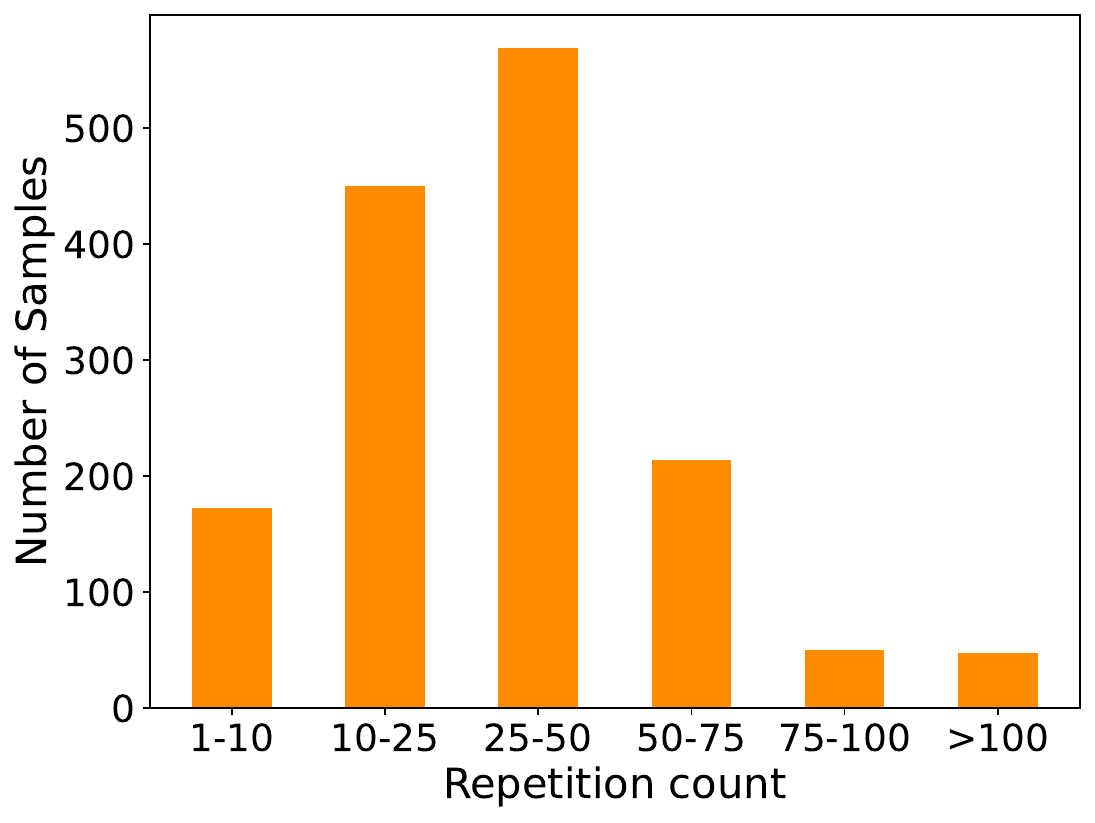}
  \end{subfigure}
  \hfill
  \begin{subfigure}[c]{0.30\textwidth}
    \centering
    \includegraphics[width=\textwidth]{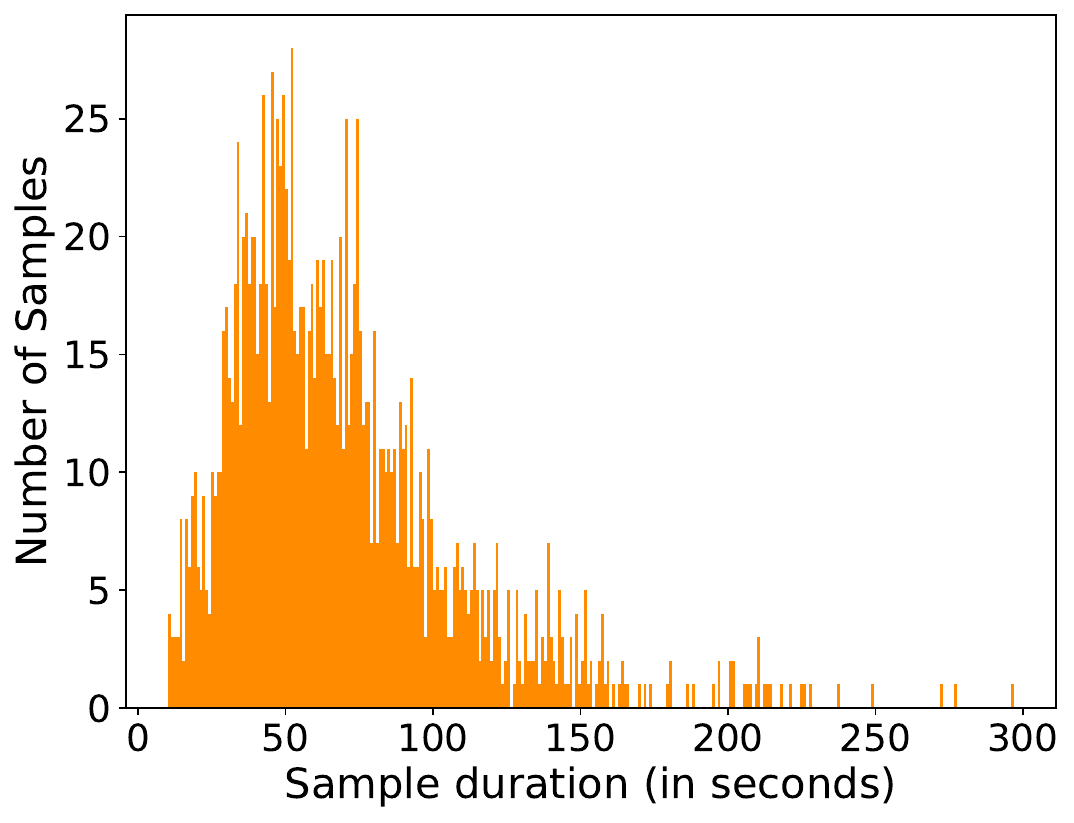}
  \end{subfigure}
  \vspace{-9px}
  \caption{\textbf{DWC dataset's statistics}: The left figure displays the action categories and the proportion of samples for each category in DWC. The two rightmost figures show the number of samples in various ranges of repetition count and duration.}
  \label{fig:dataset}
  \vspace{-20px}
\end{figure*}

\subsection{Pretraining with Synthesis Data}

Given the difficulty of collecting data from wearable devices, the amount of training data will always be limited, and it is possible that the model may overfit to the training set and subsequently underperform when faced with out-of-distribution samples. To address the issue of dataset scarcity, we propose a data synthesis method. This approach leverages the predefined vocal sounds  we previously discussed in the exemplar extraction section, effectively augmenting our existing dataset to bolster the model's robustness and ability to generalize. 
Our data synthesis approach consists of two main steps.  Firstly, we mine action templates from an existing training set. Secondly, we randomly select a template and construct a sequence by aggregating multiple, randomly augmented versions of this template, interspersed with noise or repetitive irrelevant actions.

\myheading{Action template mining}.
In the exemplar extraction, we obtain the temporal positions $i^*, j^*, k^*$ of the predefined utterances in the feature embedding sequence. We then remap these indexes to the time indexes $\hat{i}, \hat{j}, \hat{k}$ of the original sensor data sequence. Different from that, we retain the position with the minimum classification score during data synthesis. We consider $\mathcal{X}[{\hat{i}{:}\hat{j}}]$ and $\mathcal{X}[{\hat{j}{:}\hat{k}}]$ as two template candidates. We retain a candidate if it satisfies the following criteria: (1) Strong Confidence: the classification score of the temporal position greater than 0.75. This threshold ensures that we only select templates with a high degree of certainty, thus avoiding ambiguous cases. (2) Moderate Length: we discard template candidates that fall outside the established length bounds, thus avoiding excessively short or long templates that may not represent typical actions. By iterating through all the samples in the original training data, we construct an action template database, which serves as a foundation for synthesizing additional training data.

\myheading{Action sequence generation with template}.
To synthesize a training sample, we first randomly sample one action template. Then we sample the count uniformly in the range $[0.8C_l, 1.2C_u]$, where $C_l$ and $C_u$ represent the minimum and maximum counts within the training set, respectively. Afterward, we aggregate $c$ templates, augmenting each one through the following procedures: {(1) duration scaling}: we stretch or compress the duration of the template with scaling factor between 0.75 and 1.33; 
{(2) time shifting}: we shift the temporal position of the stretched/compressed template by random value within between -10 and 10 time steps; {(3) Amplitude scaling}: we modify the amplitude of the template by a scaling factor randomly chosen between 0.75 and 1.33; and {(4) random noise addition}: we introduce Gaussian noise with a standard deviation randomly chosen from 0 to 0.2. 

Through these procedures, we ensure that each synthesized training sample embodies a diversity of temporal characteristics and amplitude variations, thus enriching the synthesized training sample. Upon aggregating $c$ templates, we incorporate one to two irrelevant action sequences (described earlier) or static noise into the training sample. This integration is performed to mimic real-world data conditions, ensuring that our synthesized training data encapsulates a range of possible scenarios. 

\section{The DWC dataset}
\label{sec:dataset}

Existing datasets for action counting from wearable devices~\cite{6855613, Nishino2022FewShotAW, Zelman2020AccelerometerBasedAC, DBLP:journals/sensors/SoroBTW19, s20174791, 10.1145/3432701} often lack diversity in terms of both count values and action categories. Additionally, each data sample from these datasets also lacks diversity in terms of the actions contained within the sample, with the actions of interest being the predominant class. Considering these limitations, we introduce a more diverse dataset named DWC, which stands for Diverse and Wearable Counting. This dataset comprises 1502 entries of wearable-device data from 37 subjects across seven broad categories: kitchen activities, household chores, physical exercises, factory activities, daily routines, instrument-involved activities, and rehabilitation training. These broad categories encompass 50 distinct action classes, offering higher diversity compared to existing datasets.


We used a Samsung Galaxy Watch 4 for data collection. The sampling frequency was 100 Hz for both the 3-axis accelerometer and the 3-axis gyroscope, while the audio frequency was 16KHz. A total of 37 subjects were asked to wear the watch on their preferred hand while performing activities. Subjects were provided with a list of activities to perform in their chosen order. Each activity was accompanied by an illustrative guide and a brief textual description. The subjects were instructed to sequentially utter the words ``one,'' ``two,'' ``three'' while executing the first three repetitions of the action, with each utterance corresponding to one repetition. During data collection, participants could perform other types of action or take intermittent breaks. We manually inspected the collected data and annotated each sample with the number of repetitions of the action of interest. We also discarded samples in which the sensor and audio signals were not synchronized within 30ms. We developed an Android application to initiate the recording of both processes simultaneously, but since the audio stream was controlled by a third-party program, there were still instances of temporal mismatch.

The data was collected in two phases. In the first phase, 31 subjects participated, and each subject was asked to perform each of the 50 actions once. However, some subjects were not able to perform certain actions, such as push-ups, sit-ups, or jumping rope. The data collected in this phase containing 1356 entries with the action of interest occupying from 50\% to 90\% of the temporal duration. Upon completing the first phase, we recognized that the collected data did not possess sufficient diversity to address various practical scenarios that require counting non-dominant actions. Consequently, we proceeded with a second phase involving six additional subjects. We reviewed the list of 50 actions from the first phase and identified action classes that may not represent the predominant actions in realistic situations. Specifically, we selected six actions: picking up, shaking the clothes, slicing, tennis racket swinging, drinking and eating, and stretching. Each subject in the second phase was requested to perform each activity five times, although in some cases it was not feasible due to the lack of appropriate equipment. The data collected during this phase consists of 146 entries. These entries encompass more challenging samples where the action of interest constitutes a significantly smaller proportion of the temporal duration, ranging from only 10\% to 20\%.

The final DWC dataset consists of 1502 entries, totaling 49,258 repetitions. On average, each sample contains approximately 32 repetitions. The repetitions for individual entries range from 3 to 210. The average duration of the samples is 68.9 seconds, accumulating to almost 29 hours of sensor and audio data. The statistics are shown in~\Fref{fig:dataset}. 

\setlength{\tabcolsep}{5pt}
\begin{table}[]\footnotesize
\centering
\begin{tabular}{lrrrr}
\toprule
\multirow{2}{*}[-1ex]{Method} & \multicolumn{2}{c}{Val Set} & \multicolumn{2}{c}{Test Set}\\ 

\cmidrule(lr){2-3} \cmidrule(lr){4-5}
& \multicolumn{1}{c}{MAE} & \multicolumn{1}{c}{RMSE} & \multicolumn{1}{c}{MAE} & \multicolumn{1}{c}{RMSE}\\

Mean & 17.18 & 21.91 & 14.80 & 17.49\\
Frequency-based & 28.10 & 45.31 & 28.65 & 45.39\\

RepNet & 11.95 & 17.33 & 10.82 & 14.75\\

TransRAC & 14.51 & 20.40 & 12.97 & 16.82\\
Proposed & \textbf{7.66} & \textbf{12.25} & \textbf{7.47} & \textbf{13.09}\\   
\bottomrule 
\end{tabular}
\caption{{\bf Experiment results on DWC.} The proposed method achieves the lowest counting errors, both in terms of MAE and RMSE. Note that the Test Set is completely disjoint from the Training Set, with no overlap in terms of subjects and action categories.}
\label{tab:res}
\vspace{-20px}
\end{table}

\section{Experiments}

\myheading{Train, validation, and test data}. We conducted experiments on the DWC dataset, using a partitioning scheme that guarantees the absence of shared subjects or action categories between the training and testing data. We first divided the data into two parts, containing 35 and 15 action categories, respectively. Within each part, we further separated the subjects into two groups, one containing 25 subjects and the other 12. The combination of the 35 action categories with 25 subjects became the training set, the 15 action categories with 12 subjects formed the test set, and the remaining data constituted the validation set.



\myheading{Baselines.} We compared the proposed method against four baseline models. {\it Mean} was a method that always outputted the mean count of the samples in the training data. {\it Frequency-based} was a method that predicted the final count based on the estimated the dominant frequency. We also compared with two state-of-the-art repetitive action counting methods, namely {\it RepNet}~\cite{DBLP:conf/cvpr/DwibediATSZ20} and {\it TransRAC}~\cite{DBLP:conf/cvpr/HuDZLLG22}. To adapt these two methods for sensor data, we employed state-of-the-art feature extractors~\cite{wu2021autoformer, zhou2021informer, liu2021pyraformer} that were based on time-series forecasting and transfomers. 



\myheading{Evaluation metrics.} Following almost all previous counting methods (e.g.,~\citet{DBLP:conf/cvpr/HuDZLLG22, DBLP:conf/cvpr/Zhang0S21, DBLP:conf/iccv/LevyW15, DBLP:conf/cvpr/ZhangXHH20, DBLP:conf/cvpr/Zhang0S21}), we used Mean Absolute Error (MAE) and Root Mean Squared Error (RMSE) as performance metrics, which are defined as:  
$\text{MAE}=\frac{1}{n}\sum_{i=1}^n{|c_i - \hat{c_i}|}$; $\text{RMSE}=\sqrt{\frac{1}{n}\sum_{i=1}^n{(c_i - \hat{c_i})^2}}$,  where n is the number of test samples,
and $c_i$ and $\hat{c_i}$ are the predicted and ground truth counts.

\setlength{\tabcolsep}{2pt}
\begin{table}[]\footnotesize
\centering
\begin{tabular}{lccccc}
\toprule
\footnotesize Components & \multicolumn{5}{c}{Combinations}\\
\cmidrule{2-6}
\footnotesize Pretrain  & \xmark & \xmark & \xmark & \xmark & \checkmark \\
\footnotesize Dist. Preserving Loss & \xmark & \xmark & \xmark & \checkmark & \checkmark \\
\footnotesize Constrained Detection   & \xmark & \xmark & \checkmark & \checkmark & \checkmark \\
\footnotesize Similarity Estimation  & \xmark & \checkmark&  \checkmark & \checkmark & \checkmark \\
\midrule
\footnotesize MAE & 11.30 & 10.87 & 10.32 & 10.05 & \textbf{7.66}\\
\footnotesize RMSE & 16.15 & 15.23 & 14.96 & 14.72 & \textbf{12.25}\\
\bottomrule
\end{tabular}
\caption{\textbf{Contributions of individual components}}
\label{tab:aba:general}
\vspace{-20px}
\end{table}

\myheading{Implementation details.} The training of our model proceeded in two stages. In the first stage, the model was pre-trained on the synthesized data, which was ten times the volume of the actual training set, for 30 epochs using $\mathfrak{L}_{train}$ as the loss function. We utilized the Adam optimizer with a learning rate of $10^{-4}$ and a batch size of one for this pre-training. After pre-training, the model was trained on the actual training set for 30 epochs, using the same loss function, optimizer, and learning rate. The learning rate decay of 0.95 was applied at the end of each epoch.

During these two stages, the audio window classifier used in the exemplar extraction module was BC\_ResNet~\cite{DBLP:conf/interspeech/KimC0S21}, which was trained on Speech Command~\cite{DBLP:journals/corr/abs-1804-03209} data. The classifier was frozen and not updated during the training stages. In our model, all input sensor data was padded to a common length of 28,000. For the baseline models, the feature extraction process involved embedding the sensor data into per-window embeddings, which were then fed into the feature extractor. We standardized the window size to 50 for all baseline feature extractors. Each feature extractor consisted of three layers with a specified hidden dimension of 256 and 8 attention heads. After feature extraction, the sensor features were passed through an adaptive pooling layer of size 96 before entering the counting head. The resulting temporal self-similarity map estimated by the counting head was then processed by an MLP to generate the temporal density map.

For RepNet, the input sensor data was padded to have the length of 28,000. TransRAC did not require padding. All models underwent a training phase of 60 epochs using the Adam optimizer with a learning rate of $10^{-5}$. The training process was conducted with a batch size of one, and the count loss ($\mathfrak{L}_{c}$) was used as the loss function. All experiments were run on an RTX A5000 machine.

\myheading{Quantitative results}. 
\Tref{tab:res} shows a performance comparison of various methods on the DWC dataset. The findings highlight the superiority of the proposed method, consistently achieving a minimum 30\% lower MAE compared to other approaches. Notably, RepNet and TransRAC are strong baselines. For these baselines, extensive efforts were dedicated to optimizing their performance, tuning the pivotal feature extraction component of the methods, predominantly the time-series forecasting combined with a transformer architecture. In this pursuit, we explored a range of transformer variants, including the original transformer, Autoformer~\cite{wu2021autoformer}, Informer~\cite{zhou2021informer}, and Pyraformer~\cite{liu2021pyraformer}. Specifically, the MAE values for RepNet on the test set, when using these transformer variants, are as follows: 10.82, 13.76, 11.99, 11.29, respectively. Likewise, the corresponding MAE values for TransRAC with these transformer variants are: 12.97, 14.12, 11.55, 12.99. Despite extensive efforts to tune their performance, the resulting MAE values for these methods remain at least 30\% higher than our proposed method's MAE.

\def\subFigSz{0.45\linewidth} 
\begin{figure}[t]
    \centering
        \includegraphics[width=\subFigSz]{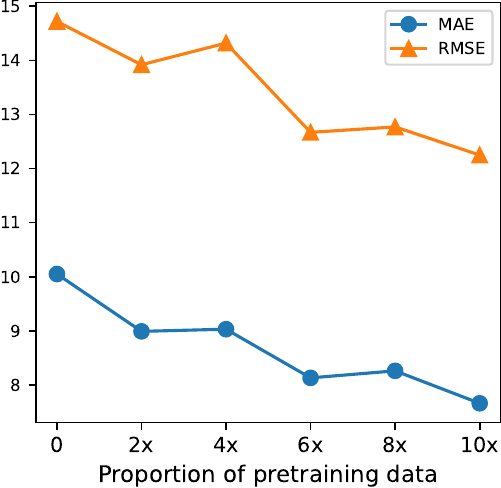}
        \includegraphics[width=\subFigSz]{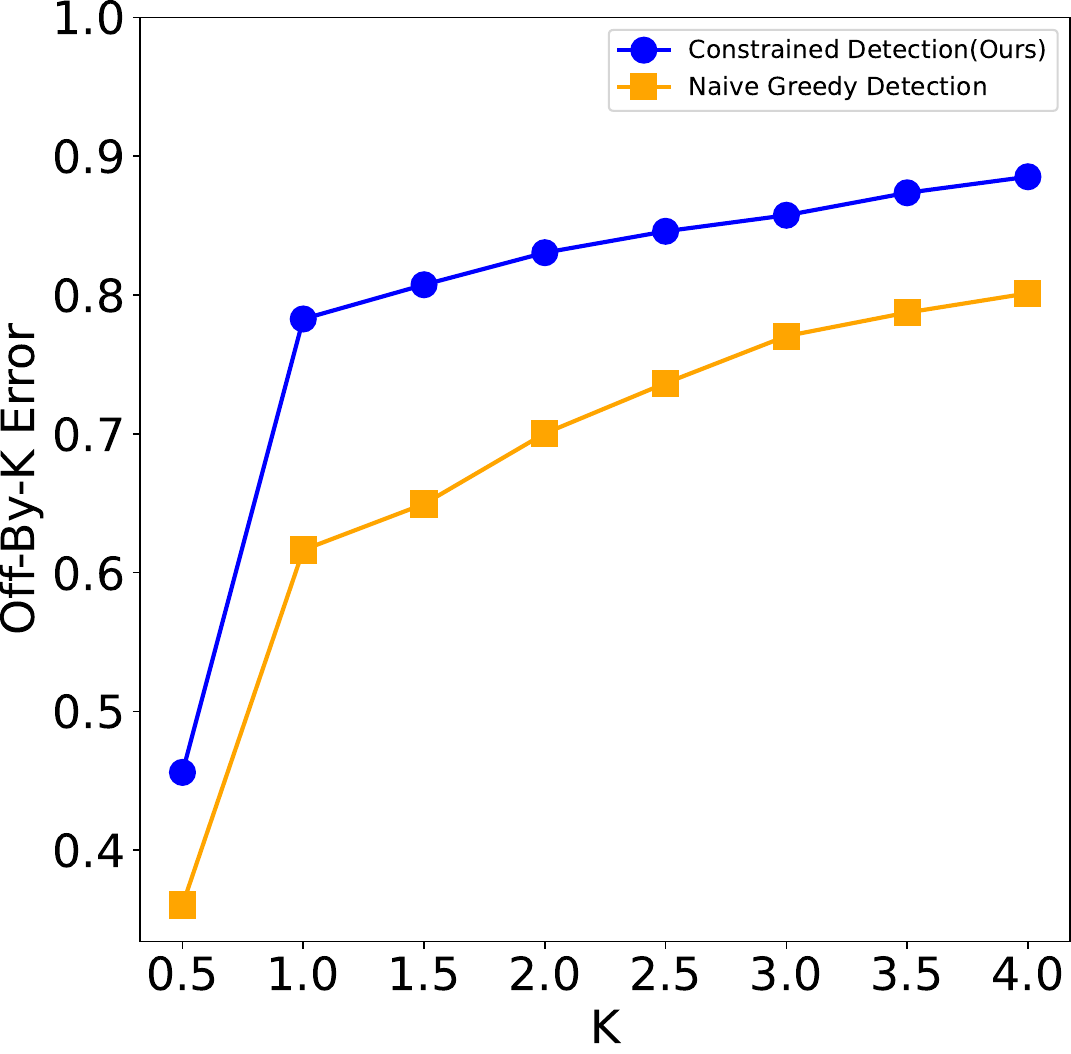} 
        \caption{\footnotesize{Left: model's performance as the amount of pretraining data is increased; ``2x'' represents twice the size of the real training set. Right: Quantitative result on temporal location detection. Off-By-K error under varying K. }}
        \label{fig:pretrain-quan}
\vspace{-18 px}
\end{figure}

\myheading{Ablation studies}. To assess the effectiveness of each component in our proposed method, we conducted an ablation study using the validation data. The results of this analysis are presented in \Tref{tab:aba:general}. The evaluated components include: (1) {\it Pretraining}: Referring to pretraining on the synthesized dataset; (2) {\it Dist. Preserving Loss}: Indicating the utilization of our distance-preserving loss; (3) {\it Constrained Detection}: Representing the use of our dynamic programming algorithm to detect the temporal locations of counting utterances under the temporal ordering and temporal proximity constraints. In its absence, we would employ a naive solution that selects the audio window with the highest classification score; and (4) {\it Similarity Estimation}: Indicating the proposed method for exemplar similarity estimation. In its absence, we use a naive correlation to estimate the similarity. The results presented in \Tref{tab:aba:general} demonstrate the beneficial impact of all proposed components on the overall performance. Particularly noteworthy is the significant contribution of pretraining on the synthesized dataset, which had the most substantial effect on the final result.

Given that pretraining is the most crucial component, we conducted further analysis to examine the impact of different amounts of pretraining data. In our default setting, we adopted an aggressive strategy, incorporating a large volume of synthesized training data, which is ten times the size of the real training data. However, we wanted to investigate whether a smaller amount of synthesized data could still yield significant improvements, resulting in faster pretraining. The results of this experiment are shown in~\Fref{fig:pretrain-quan}(a), where different proportions of the default synthesized data were used (with random selection). Specifically, ``2x'' represents twice the size of the real training set, and "4x" indicates four times the size. Intriguingly, our results reveal that even a synthesized dataset only twice the size of the real training data leads to a marked improvement in performance. Additionally, we assessed the effectiveness of using a different number of exemplars, as presented in~\Tref{tab:Exemplar_Num}.

\begin{table}[t] 
\centering
\begin{tabular}{cccccc}
\toprule 
 \multicolumn{2}{c}{One exemplar} & \multicolumn{2}{c}{Two exemplars} & \multicolumn{2}{c}{Three exemplars}\\
\cmidrule(lr){1-2} \cmidrule(lr){3-4} \cmidrule(lr){5-6}
  MAE & RMSE & MAE & RMSE & MAE & RMSE\\ 
 9.08& 14.88 & 8.74 & 14.29 & 7.66 & 12.25\\
\bottomrule 
\end{tabular}
\vspace{-0.2cm}
\caption{Experiment results on the proposed DWC validation set with different numbers of audio exemplars.}
\vspace{-0.4cm}
\label{tab:Exemplar_Num}
\end{table}

\myheading{Quantitative analysis for exemplar localization.}
Our approach relies heavily on the temporal localization of the predefined utterances. To evaluate its efficacy, we conducted an experiment on the validation set, and the result is shown in~\Fref{fig:pretrain-quan}. For evaluation, we used the Off-By-K Error (OBK) metric, defined as: $\text{OBK} = \frac{1}{N}\sum_{i=1}^N\delta(|t_i - \hat{t_i}| \leq K)$. Here, $\delta$ is the Diract delta function, $N$ represents the total number of temporal locations, $t_i$ is the predicted temporal location, and $\hat{t_i}$ is the ground truth temporal location. This metric measures the temporal discrepancy in seconds, between a predicted  location and its corresponding ground truth location. We set a naive greedy scheme as the baseline for comparison. The results of our experiment underscored the effectiveness of our approach.

\myheading{Qualitative results}. 
Qualitative results shown in \Fref{fig:QUAL} demonstrate our method's ability to accurately leverage the exemplars for counting the actions of interest and to produce a reasonable temporal density map. More qualitative results will be shown in supplementary.

\def\subFigSz{0.49\linewidth} 
\begin{figure}[] 
\centering
    \includegraphics[width=0.49\linewidth]{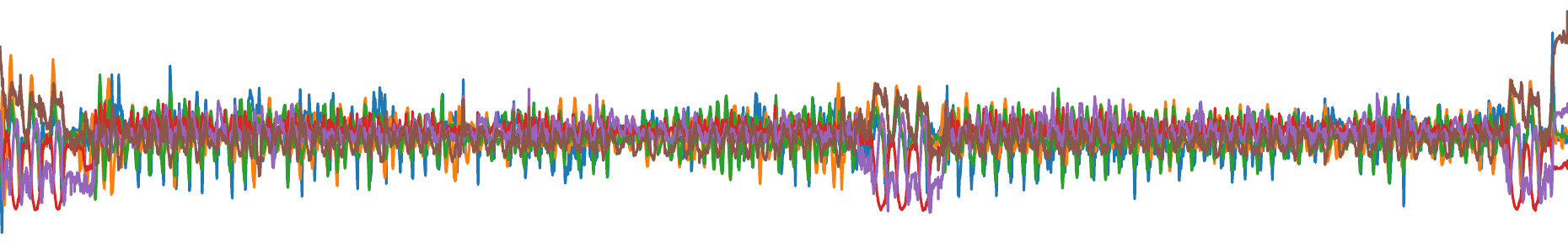}\hfill
    \includegraphics[width=\subFigSz]{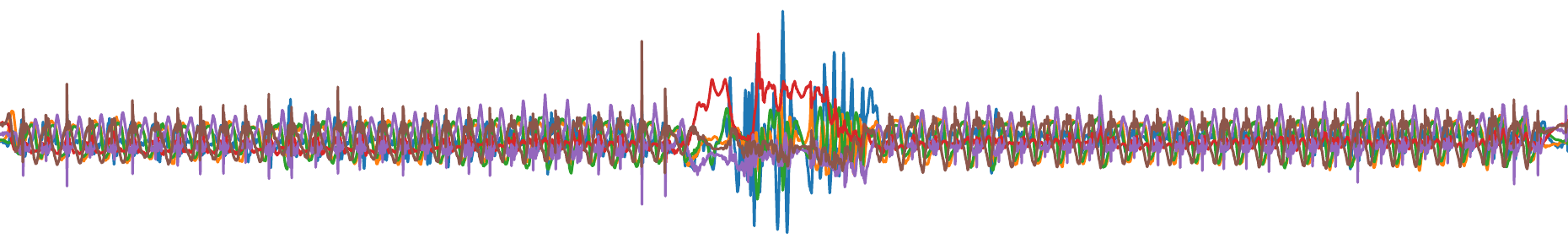}\\ 
    \includegraphics[width=0.49\linewidth]{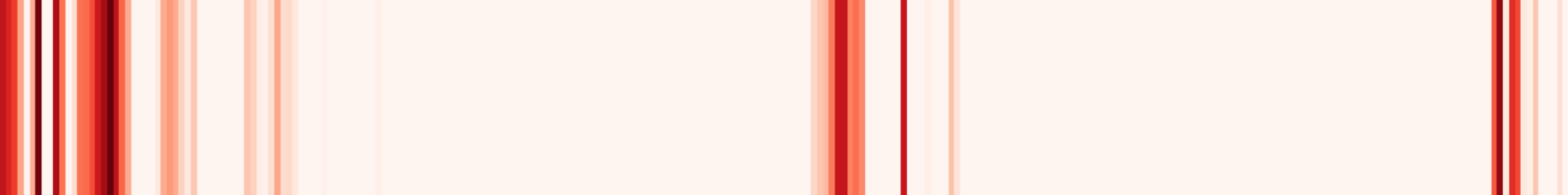}\hfill
    \includegraphics[width=\subFigSz]{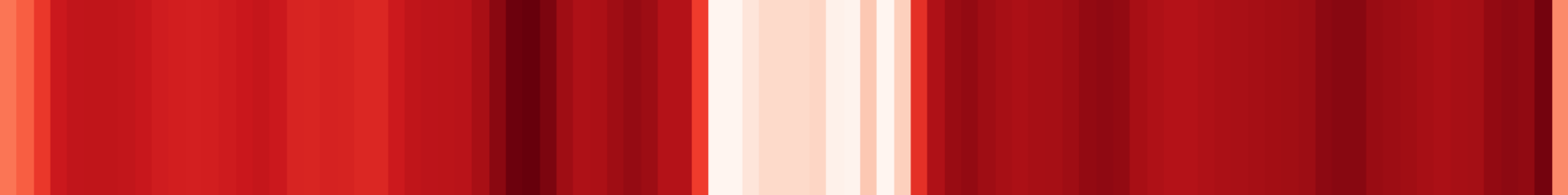}\\ 
    \makebox[0.49\linewidth]{(a) Predict: \textbf{9.7}, GT: \textbf{8}} \hfill
    \makebox[0.49\linewidth]{(b) Predict: \textbf{60.7}, GT: \textbf{60}}\\ 
    \includegraphics[width=0.49\linewidth]{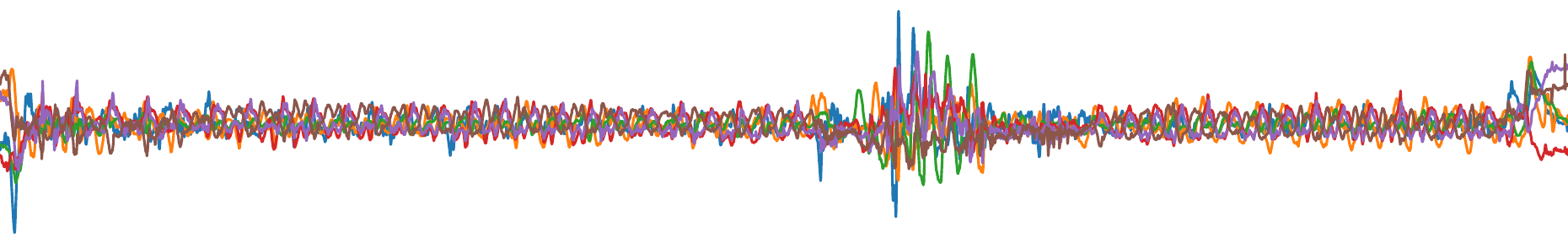}\hfill
    \includegraphics[width=\subFigSz]{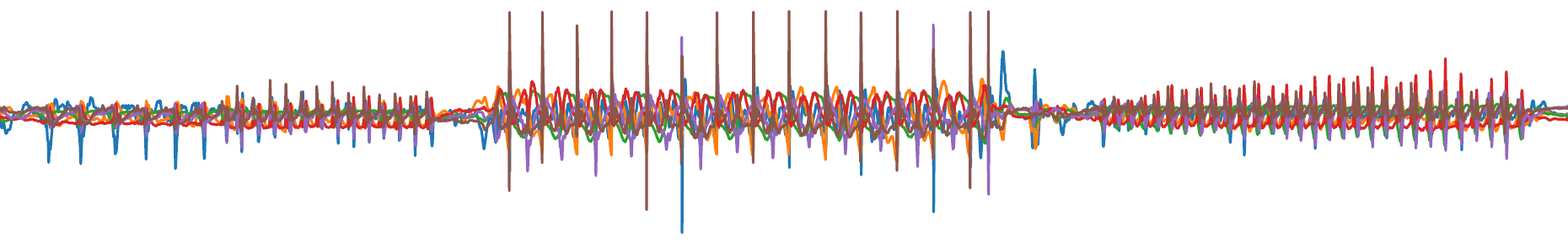}\\ 
    \includegraphics[width=0.49\linewidth]{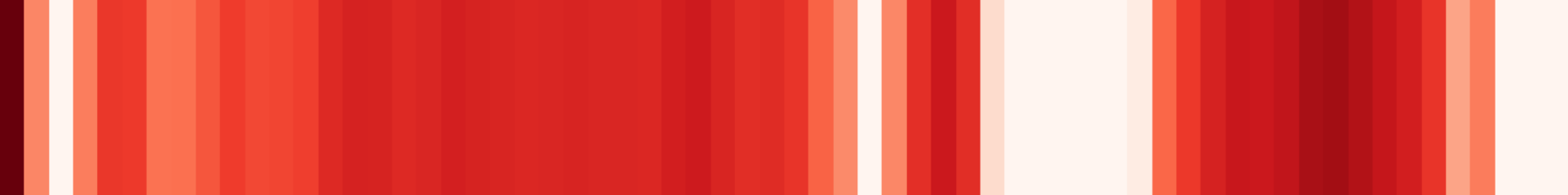}\hfill
    \includegraphics[width=\subFigSz]{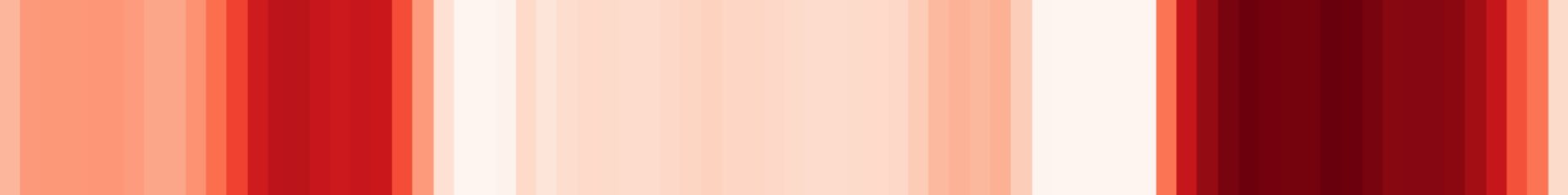}\\ 
    \makebox[0.49\linewidth]{(c) Predict: \textbf{40.7}, GT: \textbf{40}} \hfill
    \makebox[0.49\linewidth]{(d) Predict: \textbf{48.0}, GT: \textbf{50}}\\

\caption{ \textbf{Qualitative results.} Four prediction examples. Each example shows the input sensor data, the estimated density map, the predicted count, and the ground truth value.}
\label{fig:QUAL}
\vspace{-22px}
\end{figure}

\section{Conclusions}
We have proposed a few-shot approach for counting actions of interest in real-world scenarios. Our proposed approach offers a streamlined process for acquiring exemplars by detecting predetermined vocal sounds present in audio data. Furthermore, we have devised an efficient methodology to leverage these exemplars to accurately estimate temporal density values. The development of our approach have been facilitated by the introduction of an expansive and practical dataset. This dataset incorporates real-world data collected from 37 subjects across 50 distinct action categories. Experimental evaluations conducted on this dataset have demonstrated that the proposed method yields low counting errors, even for novel action classes performed by subjects not encountered in the training data.

{\myheading{Acknowledgement}.  This project was partially supported by US National Science Foundation Award NSDF DUE-2055406 and AFOSR 
Award FA2386-23-1-4058.}

\bibliography{aaai24}

\end{document}